\documentclass{article}
\usepackage{ijcai19}

\usepackage{times}
\usepackage{soul}
\usepackage{url}
\usepackage[hidelinks]{hyperref}
\usepackage[utf8]{inputenc}
\usepackage[small]{caption}
\usepackage{graphicx}
\usepackage{amsmath}
\usepackage{booktabs}
\usepackage{algorithm}
\usepackage{multirow}
\usepackage{subfig}
\usepackage{bm}
\usepackage{tabularx}                             % more flexible table environment
\usepackage{amssymb}
\usepackage[noend]{algpseudocode}
\usepackage{multicol}
\newcolumntype{L}[1]{>{\raggedright\arraybackslash}m{#1}}
\newcolumntype{P}[1]{>{\centering\arraybackslash}p{#1}}

\title{Adaptive Thompson Sampling Stacks for Memory Bounded Open-Loop Planning}

\author{
Thomy Phan \and Thomas Gabor \and Robert Müller \and Christoph Roch \and Claudia Linnhoff-Popien
\affiliations
LMU Munich
\emails
\{thomy.phan, thomas.gabor, robert.mueller, christoph.roch, linnhoff\}@ifi.lmu.de
}

\begin{document}

\maketitle

\begin{abstract}
We propose \emph{Stable Yet Memory Bounded Open-Loop (SYMBOL) planning}, a general memory bounded approach to partially observable open-loop planning.
SYMBOL maintains an adaptive stack of Thompson Sampling bandits, whose size is bounded by the planning horizon and can be automatically adapted according to the underlying domain without any prior domain knowledge beyond a generative model.
We empirically test SYMBOL in four large POMDP benchmark problems to demonstrate its effectiveness and robustness w.r.t. the choice of hyperparameters and evaluate its adaptive memory consumption. We also compare its performance with other open-loop planning algorithms and POMCP.
\end{abstract}

\section{Introduction}

\emph{Partially Observable Markov Decision Processes (POMDPs)} are useful to model many real-world problems, where the actual state is unknown to a decision making agent due to limited and noisy sensors. The agent has to consider the history of its past observations and actions to maintain a belief state as a distribution of possible states \cite{kaelbling1998planning}.

Solving POMDPs exactly is computationally intractable for domains with extremely large state spaces and long planning horizons due to the \emph{curse of dimensionality}, where the space of possible belief states grows exponentially w.r.t. the number of states \cite{kaelbling1998planning}, and the \emph{curse of history}, where the number of possible histories grows exponentially w.r.t. the horizon length \cite{pineau2006anytime}.

\emph{Monte-Carlo planning} algorithms are popular approaches to decision making in large POMDPs due to breaking both curses with statistical sampling and black box simulation \cite{silver2010monte,somani2013despot}. State-of-the-art algorithms like POMCP construct sparse \emph{closed-loop} search trees over belief states and actions. Although these approaches are computationally efficient, the search trees can become arbitrarily large in highly complex domains. In such domains, performance could be limited by restricted memory resources, which is common in sensor networks or IoT settings with intelligent devices that have to make decisions with very limited resources and perception capabilities.

\emph{Open-loop planning} is an alternative approach to efficient decision making, which only optimizes sequences of actions independently of the belief state space. Although open-loop planning generally converges to suboptimal solutions, it has been shown to be competitive against closed-loop planning in practice, when the problem is too large to provide sufficient computational resources \cite{weinstein2013open,perez2015open,lecarpentier2018open}. However, open-loop approaches have been rarely used in POMDPs so far, despite their potential to efficient planning in very large domains \cite{yu2005open,phan2019posts}.

In this paper, we propose \emph{Stable Yet Memory Bounded Open-Loop (SYMBOL) planning}, a general memory bounded approach to partially observable open-loop planning.
SYMBOL maintains an adaptive stack of Thompson Sampling bandits, which is matched to successive time steps of the decision process. The stack size is bounded by the planning horizon and can be automatically adapted on demand according to the underlying domain without any prior domain knowledge beyond a generative model.

We empirically test SYMBOL in four large benchmark problems to demonstrate its effectiveness and robustness w.r.t. the choice of hyperparameters and evaluate its adaptive memory consumption. We also compare its performance with other open-loop planning algorithms and POMCP.

\section{Background}\label{sec:background}

\subsection{POMDPs}
A POMDP is defined by a tuple $M = \langle\mathcal{S},\mathcal{A},\mathcal{P},\mathcal{R},\mathcal{O},\Omega, b_{0}\rangle$, where $\mathcal{S}$ is a (finite) set of states, $\mathcal{A}$ is the (finite) set of actions, $\mathcal{P}(s_{t+1}|s_{t}, a_{t})$ is the transition probability function, $r_{t} = \mathcal{R}(s_{t}, a_{t}) \in \mathbb{R}$ is the reward function, $\mathcal{O}$ is a (finite) set of observations, $\Omega(o_{t+1}|s_{t+1}, a_{t})$ is the observation probability function, and $b_{0}$ is a probabilitiy distribution over initial states $s_{0} \in \mathcal{S}$ \cite{kaelbling1998planning}. It is always assumed that $s_{t}, s_{t+1} \in \mathcal{S}$, $a_{t} \in \mathcal{A}$, and $o_{t}, o_{t+1} \in \mathcal{O}$ at time step $t$.

A \emph{history} $h_{t} = \big[a_{0},o_{1},...,o_{t-1},a_{t-1},o_{t}\big]$ is a sequence of actions and observations. A \emph{belief state} $b_{h_{t}}(s_{t})$ is a sufficient statistic for history $h_{t}$ and defines a probability distribution over states $s_{t}$ given $h_{t}$. $\mathcal{B}$ is the space of all possible belief states. The belief state can be updated by Bayes theorem $b_{h_{t}}(s_{t}) = \eta \Omega(o_{t}|s_{t}, a) \sum_{s \in \mathcal{S}}^{} \mathcal{P}(s_{t}|s, a) b_{h}(s)$,
where $\eta = \frac{1}{\Omega(o_{t+1}|b_{h}, a)}$ is a normalizing constant, $a = a_{t-1}$ is the last action, and $h = h_{t-1}$ is the history without $a$ and $o_{t}$.

The goal is to find a policy $\pi : \mathcal{B} \rightarrow \mathcal{A}$, which maximizes the expectation of return $G_{t}$ for a horizon $T$:
\begin{equation}\label{eq:return}
G_{t} = \sum_{k=0}^{T-1} \gamma^{k} \cdot \mathcal{R}(s_{t+k}, a_{t+k})
\end{equation}
where $\gamma \in [0,1]$ is the discount factor. %If $\gamma < 1$, then present rewards are weighted more than future rewards.

$\pi$ can be evaluated with a value function $V^{\pi}(b_{h_{t}}) = \mathbb{E}_{\pi}[G_{t}|b_{h_{t}}\big]$, which is the expected return conditioned on belief states. An optimal policy $\pi^{*}$ has a value function $V^{*}$, where $V^{*}(b_{h_{t}}) \geq V^{\pi'}(b_{h_{t}})$ for all $b_{h_{t}} \in \mathcal{B}$ and all $\pi' \neq \pi^{*}$.

\subsection{Multi-armed Bandits}
\emph{Multi-armed Bandits (MABs)} are decision making problems with a single state $s$. An agent has to repeatedly select an action $a \in \mathcal{A}$ in order to maximize its expected reward $\mathbb{E}\big[\mathcal{R}(s, a) = X_{a}]$, where $X_{a}$ is a random variable with an unknown distribution $f_{X_{a}}(x)$. The agent has to balance between exploring actions to estimate their expected reward and exploiting its knowledge on all actions by selecting the action with the currently highest expected reward. This is the \emph{exploration-exploitation dilemma}, where exploration can lead to actions with possibly higher rewards but requires time for trying them out, while exploitation can lead to fast convergence but possibly gets stuck in a local optimum. UCB1 and Thompson Sampling are possible approaches to solve MABs.

\paragraph{UCB1}
selects actions by maximizing the upper confidence bound of action values $\textit{UCB1}(a) = \overline{X_{a}} + c \sqrt{\frac{\textit{log}(n_{\textit{total}})}{n_{a}}}$, where $\overline{X_{a}}$ is the average reward of action $a$, $c$ is an exploration constant, $n_{\textit{total}}$ is the total number of action selections, and $n_{a}$ is the number of times action $a$ was selected. The second term is the \emph{exploration bonus}, which becomes smaller with increasing $n_{a}$ \cite{auer2002finite,kocsis2006bandit}.

UCB1 is a popular MAB algorithm and widely used in various challenging domains \cite{kocsis2006bandit,bubeck2010open,silver2017mastering}.

\paragraph{Thompson Sampling}
is a Bayesian approach to balance between exploration and exploitation of actions \cite{thompson1933likelihood}. The unknown reward distribution of $X_{a}$ of each action $a \in \mathcal{A}$ is modeled by a parametrized likelihood function $P_{a}(x|\theta)$ with parameter vector $\theta$. Given a prior distribution $P_{a}(\theta)$ and a set of past observed rewards $D_{a} = \{x_{1},x_{2},...,x_{n_{a}}\}$, the posterior distribution $P_{a}(\theta|D_{a})$ can be inferred by using Bayes rule $P_{a}(\theta|D_{a}) \propto \prod_{i}^{}P_{a}(x_{i}|\theta)P_{a}(\theta)$. The expected reward of each action $a \in \mathcal{A}$ can be estimated by sampling $\theta \sim P_{a}(\theta|D_{a})$ to compute $\mathbb{E}_{\theta}\big[X_{a}\big]$. The action with the highest expected reward $\mathbb{E}_{\theta}\big[X_{a}\big]$ is selected.

Thompson Sampling has been shown to be an effective and robust algorithm for making decisions under uncertainty \cite{chapelle2011empirical,kaufmann2012thompson}.

\subsection{Online Planning in POMDPs}
\emph{Planning} searches for an (near-)optimal policy given a model $\hat{M}$ of the environment $M$, which usually consists of explicit probability distributions of the POMDP. Unlike \emph{global planning}, which searches the whole (belief) state space to find an optimal policy $\pi^{*}$, \emph{local planning} only focuses on finding a policy $\pi_{t}$ for the current (belief) state by taking possible future (belief) states into account \cite{weinstein2013open}. Thus, local planning can be applied \emph{online} at every time step at the current state to recommend the next action for execution. Local planning is usually restricted to a time or computation budget $\textit{nb}$ due to strict real-time constraints \cite{bubeck2010open,weinstein2013open}.

We focus on local \emph{Monte-Carlo planning}, where $\hat{M}$ is a generative model, which can be used as black box simulator \cite{silver2010monte}. Given $s_{t}$ and $a_{t}$, the simulator $\hat{M}$ provides a sample $\langle s_{t+1},o_{t+1},r_{t}\rangle \sim \hat{M}(s_{t},a_{t})$. Monte-Carlo planning algorithms can approximate $\pi^{*}$ and $V^{*}$ by iteratively simulating and evaluating actions without reasoning about explicit probability distributions of the POMDP.

\begin{figure}[!ht]
     \subfloat[closed-loop tree\label{fig:closed_loop_planning}]{%
       \includegraphics[width=0.2\textwidth]{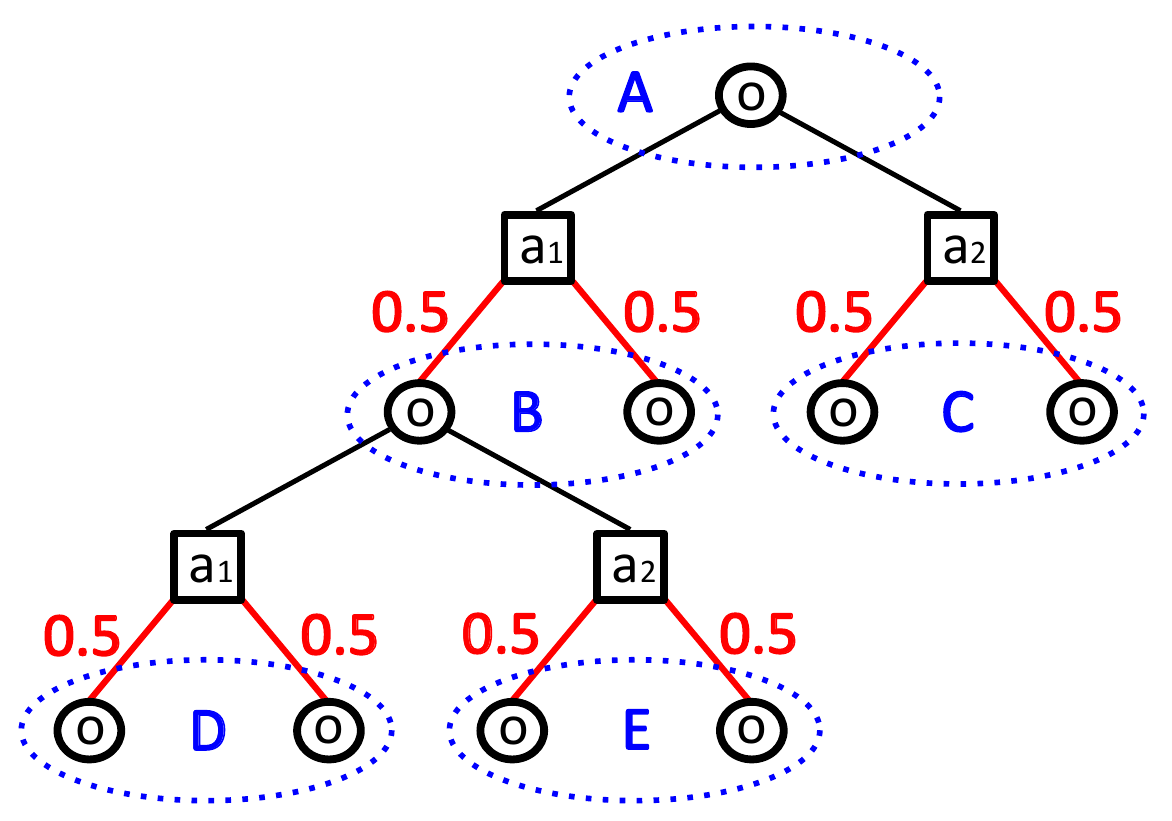}
     }
     \hfill
     \subfloat[open-loop tree\label{fig:open_loop_planning}]{%
       \includegraphics[width=0.15\textwidth]{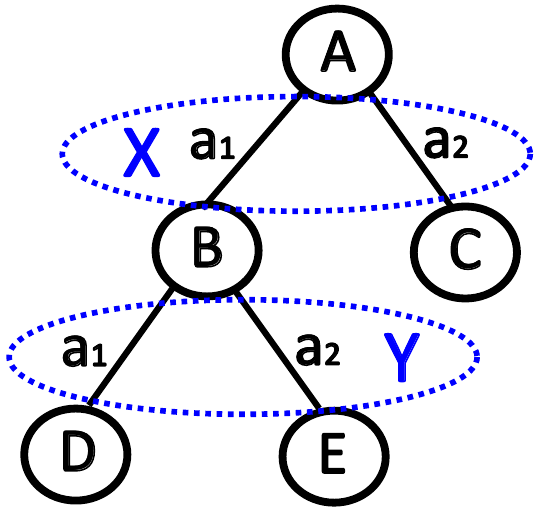}
     }\hfill
     \subfloat[stacked\label{fig:open_loop_planning_compressed}]{%
       \includegraphics[width=0.1\textwidth]{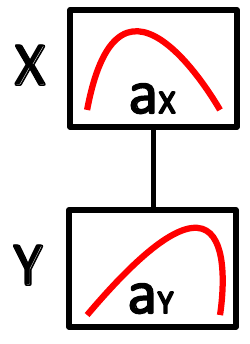}
     }
     \caption{Illustration of closed- and open-loop planning schemes. (a) Closed-loop tree with state observations (circular nodes) and actions (rectangular nodes). Red links correspond to stochastic observations made with a probability of 0.5. (b) Open-loop tree with links as actions and history distribution nodes according to the blue dotted ellipses in Fig. \ref{fig:closed_loop_planning}. (c) Open-loop approach with a stack of action distributions according to the blue dotted ellipses in Fig. \ref{fig:open_loop_planning}.}
     \label{fig:closed_vs_open_loop_planning}
\end{figure}

Local planning can be closed- or open-loop. \emph{Closed-loop planning} conditions the action selection on histories of actions and observations. \emph{Open-loop planning} only conditions the action selection on previous sequences of actions $p_{T} = [a_{1},...,a_{T}]$ (also called \emph{open-loop plans} or simply \emph{plans}) and summarized statistics about predecessor (belief) states \cite{bubeck2010open,perez2015open}. An example from \cite{phan2019posts} is shown in Fig. \ref{fig:closed_vs_open_loop_planning}. A closed-loop tree for $\Omega(o_{t+1}|s_{t},a_{t}) = 0.5$ is shown in Fig. \ref{fig:closed_loop_planning}, while Fig. \ref{fig:open_loop_planning} shows the corresponding open-loop tree which summarizes the observation nodes of Fig. \ref{fig:closed_loop_planning} within the blue dotted ellipses into \emph{history distribution nodes}. Open-loop planning can be further simplified by only regarding statistics about the expected return of actions at \emph{specific time steps} (Fig. \ref{fig:open_loop_planning_compressed}). In that case, a \emph{stack} of $T$ statistics is used to sample and evaluate plans \cite{weinstein2013open}.

\emph{Partially Observable Monte-Carlo Planning (POMCP)} is a closed-loop approach based on \emph{Monte-Carlo Tree Search (MCTS)} \cite{silver2010monte}. POMCP uses a search tree of histories with \textit{o-nodes} representing observations and \textit{a-nodes} representing actions (Fig. \ref{fig:closed_loop_planning}). The tree is traversed by selecting \textit{a-nodes} with a policy $\pi_{\textit{tree}}$ until a leaf \textit{o-node} $o_{t} \in \mathcal{O}$ is reached, which is expanded, and its value $\hat{V}(h_{t})$ is estimated with a rollout by using a policy $\pi_{\textit{rollout}}$. $\pi_{\textit{rollout}}$ can be used to integrate domain knowledge into the planning process to focus the search on promising states \cite{silver2010monte}. The observed rewards are recursively accumulated  (Eq. \ref{eq:return}) to update the value estimate of each node in the simulated path. The original version of POMCP uses UCB1 for $\pi_{\textit{tree}}$ and converges to the optimal best-first tree given sufficient computation \cite{silver2010monte}.

\cite{lecarpentier2018open} formulates an open-loop variant of MCTS using UCB1 as $\pi_{\textit{tree}}$, called \emph{Open-Loop Upper Confidence bound for Trees (OLUCT)}, which could be easily extended to POMDPs by constructing a tree, which summarizes all \textit{o-nodes} to history distribution nodes (Fig. \ref{fig:open_loop_planning}).

Open-loop planning generally converges to suboptimal solutions in stochastic domains, since it ignores (belief) state values $V(b_{h_{t}})$ and optimizes the summarized values $V(N_{t})$ of each node $N_{t}$ (Fig. \ref{fig:open_loop_planning}) instead \cite{lecarpentier2018open}. If the problem is too complex to provide sufficient computation budget $\textit{nb}$ or memory capacity, then open-loop approaches are competitive against closed-loop approaches, since they need to explore a much smaller search space to find an appropriate solution \cite{weinstein2013open,perez2015open,lecarpentier2018open}.

\section{Related Work}
Previous stack based approaches to open-loop planning maintain a fixed size stack of $T$ statistics over actions to sample open-loop plans with high expected return \cite{weinstein2013open,belzner2017stacked,phan2019posts}. While these approaches work well in practice, their convergence properties remain unclear because of the non-stationarity of the sampling statistics and the underlying state distributions due to the simultaneous adaptation of each statistic. Furthermore, the required number of statistics is highly domain dependent and hard to prespecify. SYMBOL maintains an \emph{adaptive stack} of Thompson Sampling bandits, which automatically adjusts its size according to the underlying domain without prior domain knowledge. The creation and adaptation of each bandit depends on the \emph{convergence of all preceding} bandits to preserve a stationary state and reward distribution for proper convergence of all bandits.

\cite{yu2005open} proposed an open-loop approach to decision making in POMDPs by using hierarchical planning. An open-loop plan is constructed at an abstract level, where uncertainty w.r.t. particular actions is ignored. A low-level planner controls the actual execution by explicitly dealing with uncertainty. SYMBOL is more general, since it performs planning directly on the \emph{original problem} by using a generative model for black box optimization and does not require the POMDP to be transformed for hierarchical planning.

\cite{powley2017memory} proposed a memory bounded version of MCTS with a fixed size state pool to add, discard, or reuse states depending on their visitation frequency. However, this approach cannot be easily adapted to tree-based open-loop approaches, because it requires (belief) states to be identifiable. SYMBOL does not require a pool to reuse states or nodes but maintains an \emph{adaptive stack} of Thompson Sampling bandits. The bandits adapt according to the \emph{temporal dependencies} between actions, while the size of the bandit stack is bounded by the planning horizon and \emph{automatically adapts} itself according to the underlying domain.

%\emph{Conformant Planning (CP)} is a highly related field, where states are completely unobservable and the initial state is unknown \cite{hoffmanna2006conformant,palacios2009compiling,geffner2013concise}. Since closed-loop policies are not applicable to CP problems due to unavailable observations to condition the action selection on, open-loop planning is the only way to solve such problems. CP can be regarded as a special case of POMDPs with a single uninformative observation which is observed after each time step \cite{geffner2013concise}. Since, SYMBOL is a general purpose open-loop approach to partially observable planning, which only relies on a generative model, it should be also applicable to CP problems.

\section{Adaptive Thompson Sampling Stacks}\label{sec:symbol}

\subsection{Generalized Thompson Sampling}
We use a variant of Thompson Sampling, which works for arbitrary reward distributions as proposed in \cite{bai2013bayesian,bai2014thompson} by assuming that $X_{a_{t}}$ follows a Normal distribution $\mathcal{N}(\mu, \frac{1}{\tau})$ with unknown mean $\mu$ and precision $\tau = \frac{1}{\sigma^2}$, where $\sigma^2$ is the variance. $\langle\mu, \tau\rangle$ follows a Normal Gamma distribution $\mathcal{NG}(\mu_{0},\lambda,\alpha,\beta)$ with $\lambda > 0$, $\alpha \geq 1$, and $\beta \geq 0$. The distribution over $\tau$ is a Gamma distribution $\tau \sim \textit{Gamma}(\alpha,\beta)$ and the conditional distribution over $\mu$ given $\tau$ is a Normal distribution $\mu \sim \mathcal{N}(\mu_{0},\frac{1}{\lambda\tau})$.

Given a prior distribution $P(\theta) = \mathcal{NG}(\mu_{0},\lambda_{0},\alpha_{0},\beta_{0})$ and $n$ observations $D = \{x_{1},...,x_{n}\}$, the posterior distribution is defined by $P(\theta|D) = \mathcal{NG}(\mu_{1},\lambda_{1},\alpha_{1},\beta_{1})$, where $\mu_{1} = \frac{\lambda_{0}\mu_{0} + n\overline{X}}{\lambda_{0}+n}$,
$\lambda_{1} = \lambda_{0}+n$, $\alpha_{1} = \alpha_{0}+\frac{n}{2}$, and $\beta_{1} = \beta_{0}+\frac{1}{2}(n \sigma^{2} + \frac{\lambda_{0}n(\overline{X}-\mu_{0})^2}{\lambda_{0}+n})$. $\overline{X}$ is the mean of all values in $D$ and $\sigma^{2} = \frac{1}{n}\sum_{i = 1}^{n} (x_{i} - \overline{X})^2$ is the variance.

The posterior is inferred for each action $a_{t} \in \mathcal{A}$ to sample an estimate $\mu_{a_{t}}$ for the expected return. The action with the highest $\mu_{a_{t}}$ is selected.
The complete formulation is given in Algorithm \ref{algorithm:thompson_sampling}. A MAB $N_{t}$ stores $\overline{X_{a_{t}}}$, $\sigma^{2}_{a_{t}}$, and $n_{a_{t}}$ for each action $a_{t}  \in \mathcal{A}$. In \textit{UpdateBandit}, the absolute difference $\delta_{a_{t}} = |\overline{X_{a_{t}}} - \overline{X_{\textit{old},a_{t}}}|$ between the old and the new mean value of $\overline{X_{a_{t}}}$ is returned to evaluate the convergence of $N_{t}$.

\begin{algorithm}
\caption{Generalized Thompson Sampling}\label{algorithm:thompson_sampling}
\begin{algorithmic}
\Procedure{$\textit{ThompsonSampling}(N_{t})$}{}
\For{$a_{t} \in \mathcal{A}$}
\State Infer $\langle \mu_{1},\lambda_{1},\alpha_{1},\beta_{1} \rangle$ from prior and $\overline{X_{a_{t}}},\sigma^{2}_{a_{t}}, n_{a_{t}}$
\State	$\mu_{a_{t}}, \tau_{a_{t}} \sim \mathcal{NG}(\mu_{1},\lambda_{1},\alpha_{1},\beta_{1})$
\EndFor
\Return $\textit{argmax}_{a_{t} \in \mathcal{A}}(\mu_{a_{t}})$
\EndProcedure
\end{algorithmic}
\begin{algorithmic}
\Procedure{$\textit{UpdateBandit}(N_{t}, G_{t}, a_{t})$}{}
\State $\langle \overline{X_{\textit{old},a_{t}}}, \overline{X_{a_{t}}} \rangle \leftarrow \langle\overline{X_{a_{t}}}, (n_{a_{t}}\overline{X_{\textit{old},a_{t}}} + G_{t})/(n_{a_{t}}+1)\rangle$
\State $n_{a_{t}} \leftarrow n_{a_{t}} + 1$
\State $\sigma^{2}_{a_{t}} \leftarrow [(n_{a_{t}}-1)\sigma^{2}_{a_{t}} + (G_{t} - \overline{X_{\textit{old},a_{t}}})(G_{t} - \overline{X_{a_{t}}})]/n_{a_{t}}$
\State $\delta_{a_{t}} \leftarrow |\overline{X_{a_{t}}} - \overline{X_{\textit{old},a_{t}}}|$
\State\Return $\delta_{a_{t}}$
\EndProcedure
\end{algorithmic}
\end{algorithm}

%The prior should ideally reflect knowledge about the underlying model, especially for initial turns, where only a small amount of data has been observed \cite{honda2014optimality}. If no knowledge is available, then \emph{uninformative priors} should be chosen, where all possibilities can be sampled (almost) uniformly.
If sufficient domain knowledge for defining the prior is unavailable, the prior should be chosen such that all possibilities can be sampled (almost) uniformly \cite{bai2014thompson}.
This can be achieved by choosing the priors such that the variance of the resulting Normal distribution $\mathcal{N}(\mu_{0}, \frac{1}{\lambda_{0}\tau})$ becomes infinite ($\frac{1}{\lambda_{0}\tau_{0}} \rightarrow \infty$ and $\lambda_{0}\tau \rightarrow 0$). Since $\tau$ follows a Gamma distribution $\textit{Gamma}(\alpha_{0},\beta_{0})$ with expectation $\mathbb{E}\big[\tau\big] = \frac{\alpha_{0}}{\beta_{0}}$, $\alpha_{0}$ and $\beta_{0}$ should be chosen such that $\frac{\alpha_{0}}{\beta_{0}} \rightarrow 0$. Given the hyperparameter space $\lambda_{0} > 0$, $\alpha_{0} \geq 1$, and $\beta_{0} \geq 0$, it is recommended to set $\alpha_{0} = 1$ and $\mu_{0} = 0$ to center the Normal distribution. $\lambda_{0}$ should be small enough and $\beta_{0}$ should be sufficiently large \cite{bai2014thompson}.

\subsection{SYMBOL}\label{subsec:symbol}
\emph{Stable Yet Memory Bounded Open-Loop (SYMBOL) planning} is a partially observable open-loop approach, which optimizes an adaptive stack of $\textit{nMAB}$ Thompson Sampling bandits (with $\textit{nMAB} \leq T$) to maximize the expected return.

Initially beginning with a single MAB $N_{1}$, a simulation starts at state $s_{t}$, which is sampled from an approximated belief state $\hat{b}_{h_{t}}$ \footnote{We use a particle filter for $\hat{b}_{h_{t}} \approx b_{h_{t}}$ \cite{silver2010monte}.}. The first $\textit{nMAB}$ actions $a_{t}$ are sampled from all MABs in the current stack. The remaining $T - \textit{nMAB}$ actions are sampled from a rollout policy $\pi_{\textit{rollout}}$, which can be random or enhanced with domain knowledge \cite{silver2010monte}. The sampled plan $p_{T} = [a_{1},...,a_{T}]$ is evaluated with $\hat{M}$ to observe rewards $r_{1},...,r_{T}$, which are accumulated to returns $G_{1},...,G_{T}$ according to Eq. \ref{eq:return}. The first $\textit{nMAB}$ returns are used to update the MAB stack. A MAB $N_{t}$ is only updated or created, when all of its predecessors $N_{t-k}$ with $0 < k < t$ converged. We assume that $N_{t}$ converged, if $\overline{\delta}_{a_{t}} < \epsilon$, where $\overline{\delta}_{a_{t}}$ is the average of the last $\kappa$ values of $\delta_{a_{t}}$ from previous updates to $N_{t}$ (Algorithm \ref{algorithm:thompson_sampling}). The parameter $\kappa$ copes with the non-stationarity of the return values $G_{t}$ to update each MAB, which is caused by the adaptation of the action selection.

The complete formulation of SYMBOL is given in Algorithm \ref{algorithm:symbol}, where $h_{t}$ is the action-observation history, $T$ is the planning horizon, $\textit{nb}$ is the computation budget, $\kappa$ is the \emph{convergence tolerance} represented by the number of MAB updates to be considered, and $\epsilon$ is the \emph{convergence threshold}.

\begin{algorithm}
\caption{SYMBOL Planning}\label{algorithm:symbol}
\begin{algorithmic}
\Procedure{$\textit{SYMBOL}(h_{t},T,\textit{nb},\kappa,\epsilon)$}{}
\State $\textit{nMAB} \leftarrow 1$, Create first MAB $N_{1}$
\While{$\textit{nb} > 0$}
\State $\textit{nb} \leftarrow \textit{nb} - 1$
\State $s_{1} \sim \hat{b}_{h_{t}}$
\State $\textit{Simulate}(s_{1},T,\kappa,\epsilon)$
\EndWhile
\State\Return $\textit{argmax}_{a_{1} \in \mathcal{A}}(\overline{X_{a_{1}}})$
\EndProcedure
\end{algorithmic}
\begin{algorithmic}
\Procedure{$\textit{Simulate}(s_{1},T,\kappa,\epsilon)$}{}
\State $t \leftarrow 1$
\While{$t \leq T$ and $s_{t}$ is no terminal state}
	\If{$t \leq \textit{nMAB}$}
	\State $a_{t} \leftarrow \textit{ThompsonSampling}(N_{t})$
	\Else
	\State $a_{t} \leftarrow \pi_{\textit{rollout}}(s_{t})$
	\EndIf
	\State $\langle s_{t+1}, r_{t},o_{t+1} \rangle \sim \hat{M}(s_{t}, a_{t})$ \Comment{Simulate action}
	\State $t \leftarrow t + 1$
\EndWhile
\State $\langle H, G_{t+1} \rangle \leftarrow \langle t, 0 \rangle$
\For{$t \in {H, ..., 1}$} \Comment{Accumulate rewards (Eq. \ref{eq:return})}
	\State $G_{t} \leftarrow r_{t} + \gamma G_{t+1}$
\EndFor
\For{$t \in {1, ..., H}$}
	\If{$t \leq \textit{nMAB} + 1$ and $t \leq T$ and $\overline{\delta}_{a_{t-1}} < \epsilon$}
		\If{$t > \textit{nMAB}$}
			%\State Create new $N_{t}$ \Comment{Expand MAB stack}
			\State $\textit{nMAB} \leftarrow \textit{nMAB} + 1$, Create new $N_{t}$
		\EndIf
		\State $\delta_{a_{t}} \leftarrow \textit{UpdateBandit}(N_{t}, G_{t}, a_{t})$
		\State Update $\overline{\delta}_{a_{t}}$ with average of last $\kappa$ values of $\delta_{a_{t}}$
    \Else
    	\State \textbf{break} \Comment{Keep successor MABs stationary}
	\EndIf
\EndFor
\EndProcedure
\end{algorithmic}
\end{algorithm}

Thompson Sampling bandits are able to converge, if their reward distributions are stationary \cite{agrawal2013further}. A reward distribution is stationary, if both the underlying state distribution and the successor policy $\pi_{\textit{succ}}$ are stationary. The former is ensured, if all preceding MABs converged, since their actions affect the underlying state distribution. The latter is ensured by keeping all successing MABs fixed (they are not updated unless the predecessors converged) and by using a stationary rollout policy $\pi_{\textit{rollout}}$. 

Starting from the first MAB $N_{1}$, we know that the state distribution $\hat{b}_{h_{t}}$ is stationary. If the successor policy $\pi_{\textit{succ}}$ is stationary as well, $N_{1}$ will converge to the best action given $\pi_{\textit{succ}}$ \cite{agrawal2013further}. By induction, we can show the same for all successor MABs, given that all predecessor MABs converged. If $N_{t}$ significantly changed such that $\overline{\delta}_{a_{t}} \geq \epsilon$, all MABs $N_{t+k}$ with $k > 0$ must remain fixed to ensure a stationary reward distribution to enable convergence of $N_{t}$ first.
By adjusting the parameters $\kappa$ and $\epsilon$, the size and speed of convergence of the MAB stack can be controlled.

\section{Experiments}\label{sec:experiments}
\subsection{Evaluation Environments}

We tested SYMBOL in different POMDP benchmark problems \cite{silver2010monte,somani2013despot}. We always set $\gamma$ as proposed in \cite{silver2010monte}.

The \textit{RockSample(n,k)} problem simulates an agent moving in an $n \times n$ grid, which contains $k$ rocks \cite{smith2004heuristic}. Each rock can be $good$ or $bad$, but the true state of each rock is unknown. The agent has to sample good rocks, while avoiding to sample bad rocks. It has a noisy sensor, which produces an observation $o_{t} \in \{good,bad\}$ for a particular rock. The probability of sensing the correct state of the rock decreases exponentially with the agent's distance to that rock. Sampling gives a reward of $+10$, if the rock is good, and $-10$ otherwise. If a good rock was sampled, it becomes bad. Moving past the east edge of the grid gives a reward of $+10$ and the episode terminates. We set $\gamma = 0.95$.

In \textit{Battleship} five ships of size 1, 2, 3, 4, and 5 respectively are randomly placed into a $10 \times 10$ grid, where the agent has to sink all ships without knowing their actual positions \cite{silver2010monte}. Each cell hitting a ship gives a reward of $+1$. There is a reward of $-1$ per time step and a terminal reward of $+100$ for hitting all ships. We set $\gamma = 1$.

\textit{PocMan} is a partially observable version of \textit{PacMan} \cite{silver2010monte}. The agent navigates in a $17 \times 19$ maze and has to eat randomly distributed food pellets and power pills. There are four ghosts moving randomly in the maze. If the agent is within the visible range of a ghost, it is getting chased by the ghost and dies, if it touches the ghost, terminating the episode with a reward of $-100$. Eating a power pill enables the agent to eat ghosts for 15 time steps. In that case, the ghosts will run away, if the agent is under the effect of a power pill. At each time step a reward of $-1$ is given. Eating food pellets gives a reward of $+10$ and eating a ghost gives $+25$. The agent can only perceive ghosts, if they are in its direct line of sight in each cardinal direction or within a hearing range. Also, the agent can only sense walls and food pellets, which are adjacent to it. We set $\gamma = 0.95$.

\subsection{Methods}\label{subsec:methods}

We implemented different partially observable planning algorithms to compare with SYMBOL \footnote{Code available at \url{https://github.com/thomyphan/planning}}. All algorithms with a rollout phase use a policy $\pi_{\textit{rollout}}$ which randomly selects actions from a set of legal actions $a_{t} \in \mathcal{A}_{\textit{legal}}(s')$, depending on the currently simulated state $s' \in \mathcal{S}$. Since open-loop planning can encounter different states at the same node or time step (Fig. \ref{fig:closed_vs_open_loop_planning}), the set of legal actions $\mathcal{A}_{\textit{legal}}(s_{t})$ may vary for each state $s_{t} \in \mathcal{S}$. Thus, we mask out currently illegal actions, regardless of whether they have high action values.

\paragraph{POMCP}
We use the POMCP implementation from \cite{silver2010monte}. $\pi_{\textit{tree}}$ selects actions from $\mathcal{A}_{\textit{legal}}(s_{t})$ with UCB1. In each simulation step, there is at most one expansion step, where new nodes are added to the search tree. Thus, the tree size should increase linearly w.r.t. $\textit{nb}$ in large POMDPs.

\paragraph{POOLUCT and POOLTS}
are implemented as open-loop versions of POMCP (Fig. \ref{fig:open_loop_planning}), where actions are selected from $\mathcal{A}_{\textit{legal}}(s_{t})$ using UCB1 (POOLUCT) or Thompson Sampling (POOLTS) as node selection policy $\pi_{\textit{tree}}$.
Similarly to POMCP, the search tree size should increase linearly w.r.t. $\textit{nb}$, but with less nodes, since open-loop trees store summarized information about history distributions (Fig. \ref{fig:open_loop_planning}).

\paragraph{SYMBOL}
uses an adaptive stack of $\textit{nMAB}$ Thompson Sampling bandits $N_{t}$ according to Algorithm \ref{algorithm:symbol}. Starting at $s_{1}$, all MABs $N_{t}$ apply Thompson Sampling to $\mathcal{A}_{\textit{legal}}(s')$, depending on the currently simulated state $s' \in \mathcal{S}$. If $t > \textit{nMAB}$, then $\pi_{\textit{rollout}}$ is used.
Given a horizon of $T$, SYMBOL always  maintains $\textit{nMAB} \leq T$ MABs. Although $\textit{nMAB}$ depends on $\kappa$, $\epsilon$, and $\textit{nb}$, it never exceeds $T$ (Algorithm \ref{algorithm:symbol}).

\paragraph{Partially Observable Stacked Thompson Sampling (POSTS)}
uses a \emph{fixed size stack} of $\textit{nMAB} = T$ Thompson Sampling bandits $N_{t}$ \cite{phan2019posts}. Similarly to SYMBOL, all MABs $N_{t}$ apply Thompson Sampling to $\mathcal{A}_{\textit{legal}}(s')$, depending on the currently simulated state $s' \in \mathcal{S}$. Unlike SYMBOL, all MABs $N_{t}$ are updated \emph{simultaneously} according to $G_{t}$ regardless of the convergence of the preceding MABs as suggested in \cite{weinstein2013open,belzner2017stacked,phan2019posts}.

\subsection{Results}
We ran each approach on \textit{RockSample}, \textit{Battleship}, and \textit{PocMan} with different settings for 100 times or at most 12 hours of total computation. We evaluated the performance of each approach with the \emph{undiscounted return} ($\gamma = 1$), because we focus on the actual effectiveness instead of the quality of optimization \cite{bai2014thompson}. For POMCP and POOLUCT we set the UCB1 exploration constant $c$ to the reward range of each domain as proposed in \cite{silver2010monte}.

Since we assume no additional domain knowledge, we focus on uninformative priors with $\mu_{0} = 0$, $\alpha_{0} = 1$, and $\lambda_{0} = 0.01$ \cite{bai2014thompson}. With this setting, $\beta_{0}$ controls the degree of initial exploration during the planning phase, thus we only vary $\beta_{0}$ for POOLTS, POSTS, and SYMBOL. To preserve readability of the figures, we only provide the best configuration of POOLTS and POSTS for comparison.

\subsubsection{Hyperparameter Sensitivity}
\begin{figure*}[!ht]
	\subfloat[\textit{RockSample(11,11)}\label{fig:rocksample11_epsilon_memory_performance}]{%
       \includegraphics[width=0.24\textwidth]{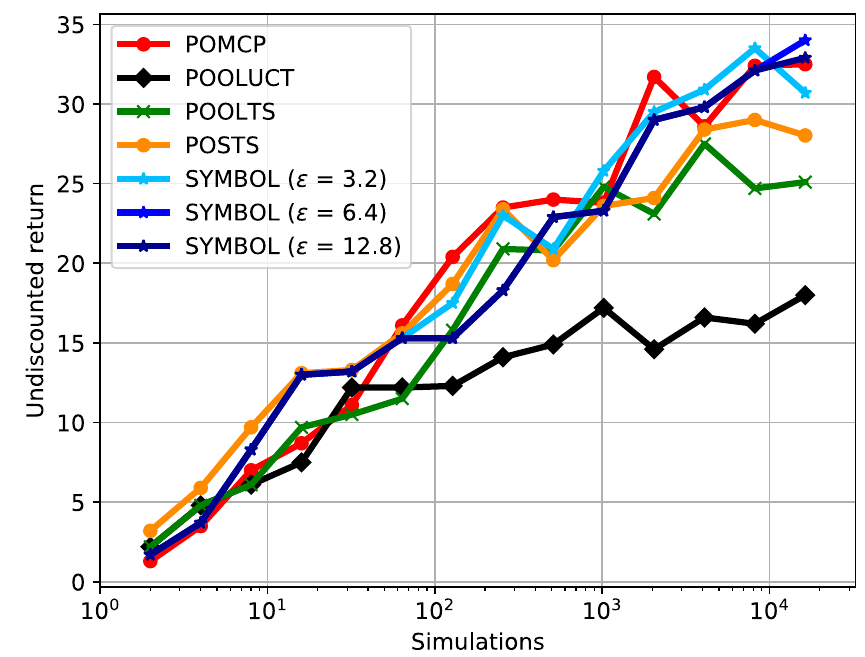}
     }
     \hfill
     \subfloat[\textit{RockSample(15,15)}\label{fig:rocksample15_epsilon_memory_performance}]{%
       \includegraphics[width=0.24\textwidth]{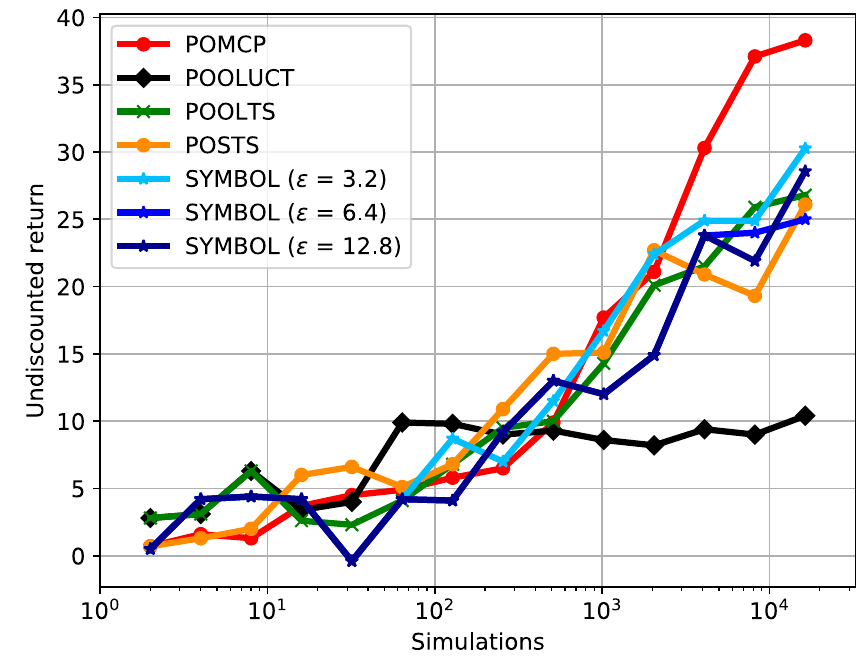}
     }
     \subfloat[\textit{Battleship}\label{fig:battleship_epsilon_memory_performance}]{%
       \includegraphics[width=0.24\textwidth]{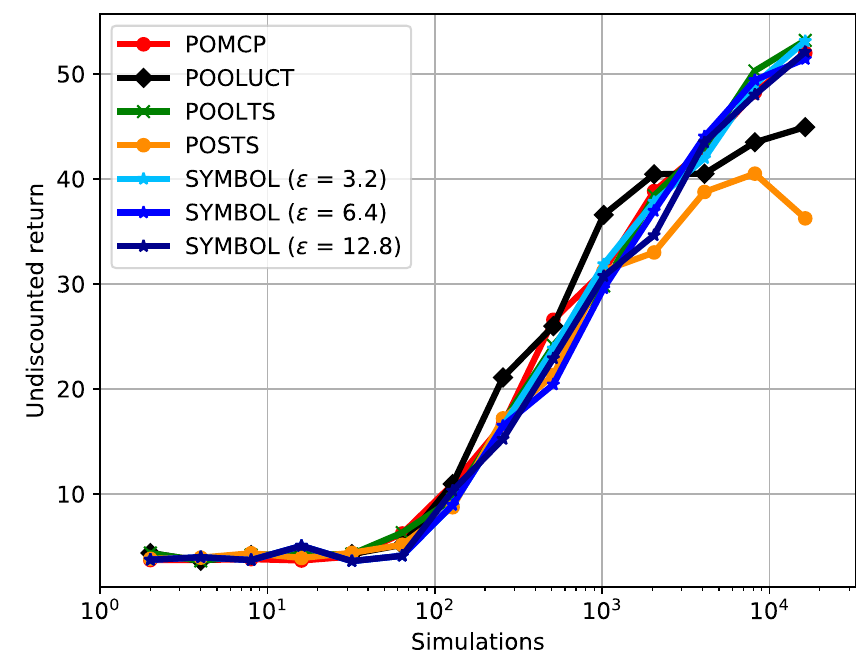}
     }
     \hfill
     \subfloat[\textit{PocMan}\label{fig:pocman_epsilon_memory_performance}]{%
       \includegraphics[width=0.24\textwidth]{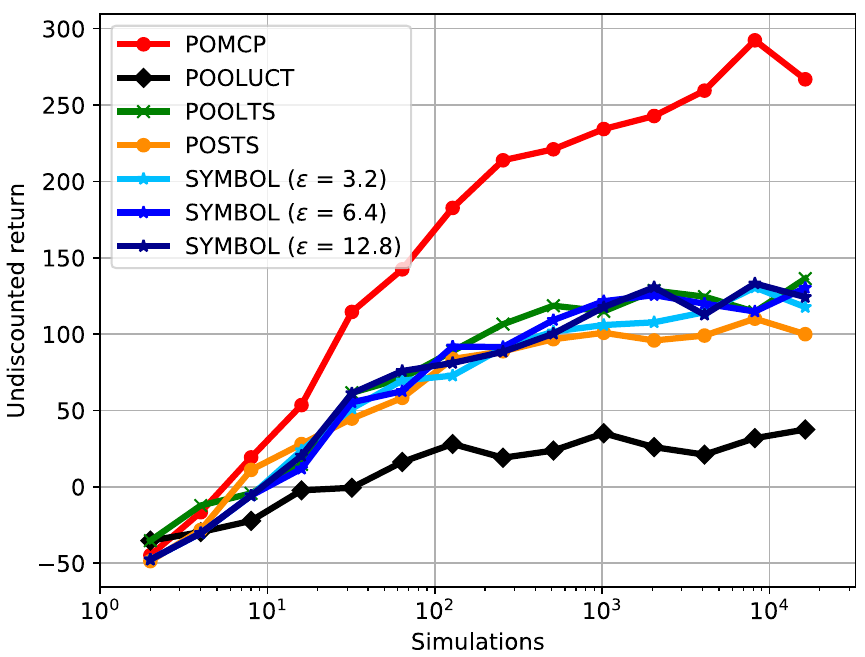}
     }
     \\
     \subfloat[\textit{RockSample(11,11)}\label{fig:rocksample11_epsilon_memory}]{%
       \includegraphics[width=0.24\textwidth]{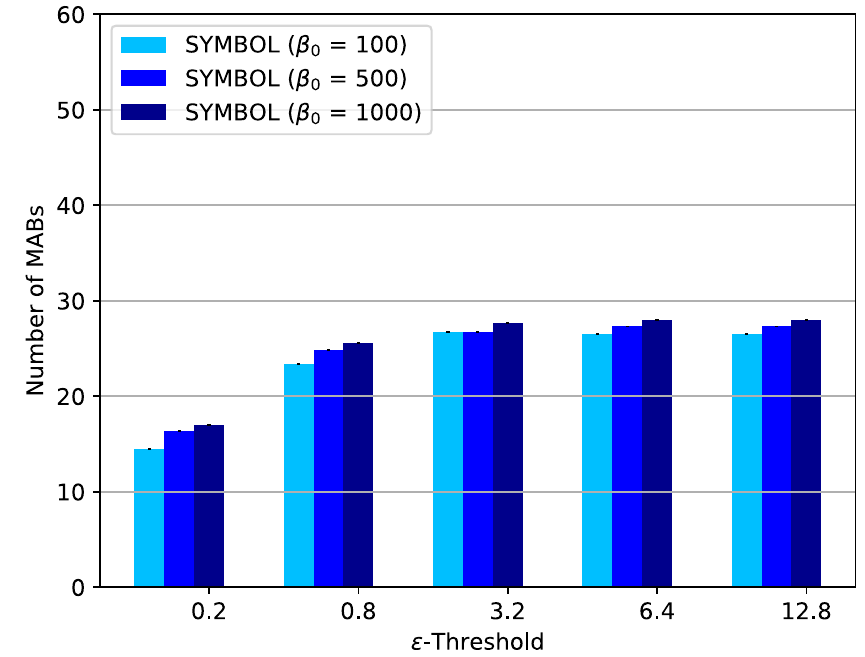}
     }
     \hfill
     \subfloat[\textit{RockSample(15,15)}\label{fig:rocksample15_epsilon_memory}]{%
       \includegraphics[width=0.24\textwidth]{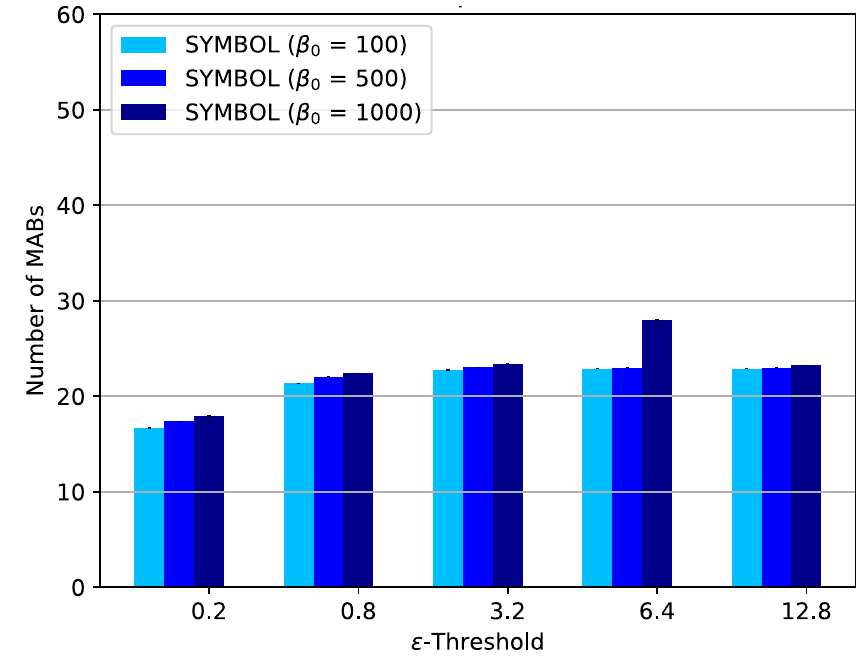}
     }
     \subfloat[\textit{Battleship}\label{fig:battleship_epsilon_memory}]{%
       \includegraphics[width=0.24\textwidth]{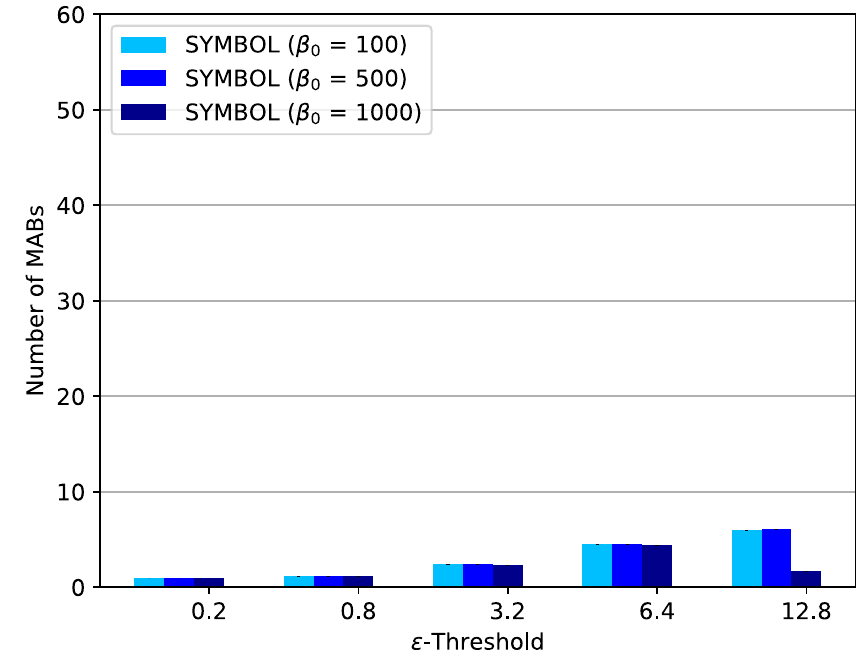}
     }
     \hfill
     \subfloat[\textit{PocMan}\label{fig:pocman_epsilon_memory}]{%
       \includegraphics[width=0.24\textwidth]{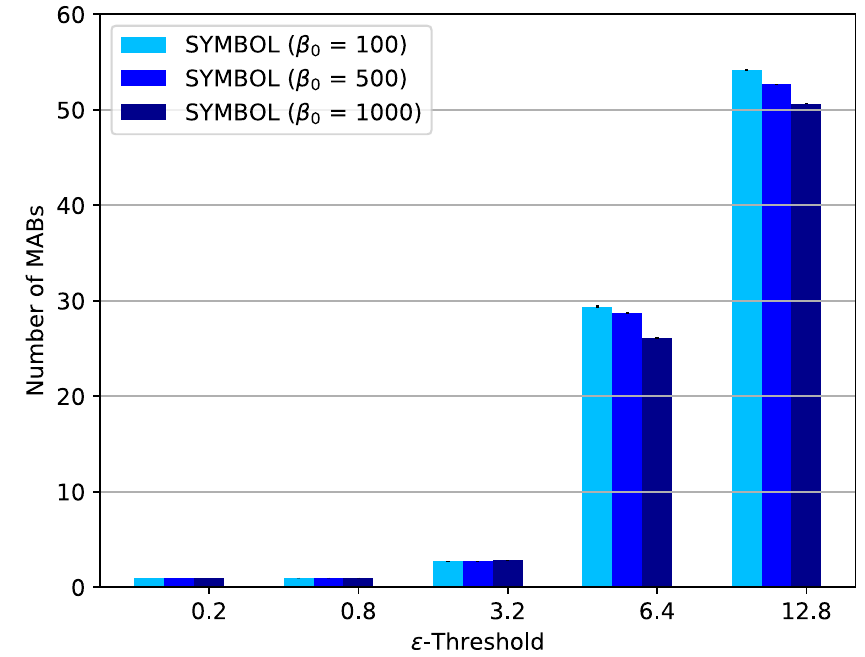}
     }
     \caption{Sensitivity analysis of SYMBOL w.r.t. the $\epsilon$-threshold with a horizon of $T = 100$ and $\kappa = 8$. (a-d) Average performance w.r.t. to different computation budgets $\textit{nb}$ compared to POMCP, POOLUCT, POOLTS, and POSTS (e-h) MAB stack size given $\textit{nb} = 4096$.}
     \label{fig:symbol_epsilon_sensitivity}
\end{figure*}

We evaluated the sensitivity of SYMBOL w.r.t. the convergence threshold $\epsilon$.
Fig. \ref{fig:symbol_epsilon_sensitivity}a-d show the performance of SYMBOL with $\epsilon \in \{3.2, 6.4, 12.8\}$ compared to all other approaches described in Section \ref{subsec:methods}. All SYMBOL variants are able to keep up with their tree-based counterpart POOLTS, while outperforming POOLUCT. SYMBOL is able to keep up with POMCP in \textit{RockSample(11,11)} and \textit{Battleship}. POMCP outperforms all open-loop approaches in \textit{PocMan}. SYMBOL scales better in performance with increasing $\textit{nb}$ than POSTS, which seems to converge prematurely after $\textit{nb} > 1000$. Except in \textit{PocMan}, SYMBOL scales slightly better with increasing $\textit{nb}$ when $\epsilon = 3.2$, probably due to more stable convergence of the MABs.
Fig. \ref{fig:symbol_epsilon_sensitivity}e-h show the average stack sizes $\textit{nMAB}$ of SYMBOL for $\textit{nb} = 4096$ \footnote{Using budgets between 1024 and 16384 led to similar plots, thus we stick to $\textit{nb} = 4096$ as suggested in \cite{phan2019posts}.}, $\beta_{0} \in \{100, 500, 1000\}$, and different $\epsilon$. In \textit{RockSample}, $20 < \textit{nMAB} < 30$, when $\epsilon \geq 0.8$, but it does not grow any further. In \textit{Battleship}, $\textit{nMAB} < 10$, but the stack size slightly increases w.r.t $\epsilon$ . In \textit{PocMan}, $\textit{nMAB}$ quickly increases w.r.t $\epsilon$. $\beta_{0}$ does not have any significant impact on $\textit{nMAB}$.

We also experimented with the convergence tolerance $\kappa \in \{2, 4, 8, 16, 32\}$ but did not observe significantly different results than shown in Fig. \ref{fig:symbol_epsilon_sensitivity}. $\textit{nMAB}$ tends to decrease with increasing $\kappa$, which is due to the amount of time required to consider a MAB as converged. When $\kappa < 8$, then SYMBOL was less stable in all domains (except in \textit{Battleship}), leading to high variance in performance when $\textit{nb}$ is large.

\subsubsection{Performance-Memory Tradeoff}
\begin{figure}[!ht]
     \subfloat[\textit{RockSample(11,11)}\label{fig:rocksample_11_memory}]{%
       \includegraphics[width=0.23\textwidth]{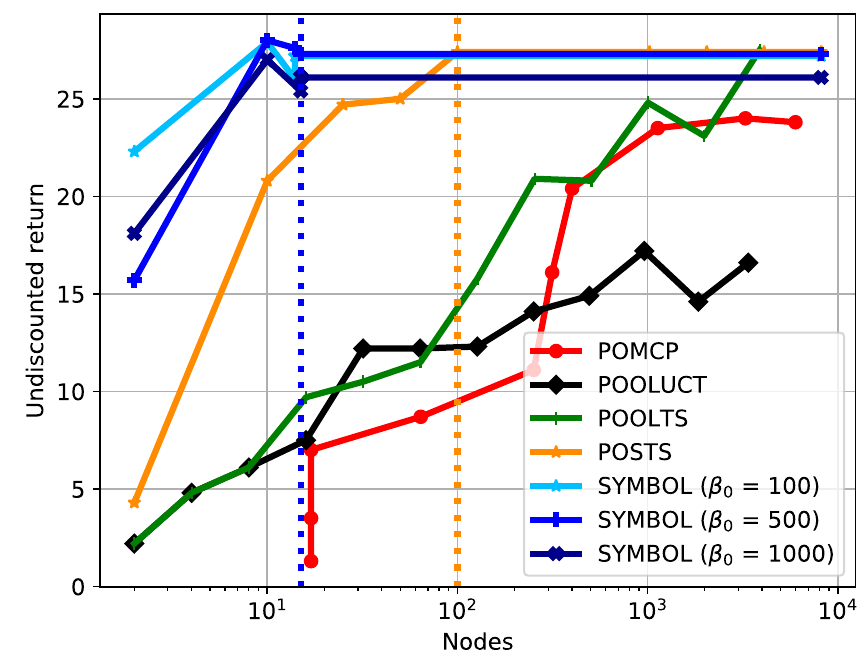}
     }
     \hfill
     \subfloat[\textit{RockSample(15,15)}\label{fig:rocksample_15_memory}]{%
       \includegraphics[width=0.23\textwidth]{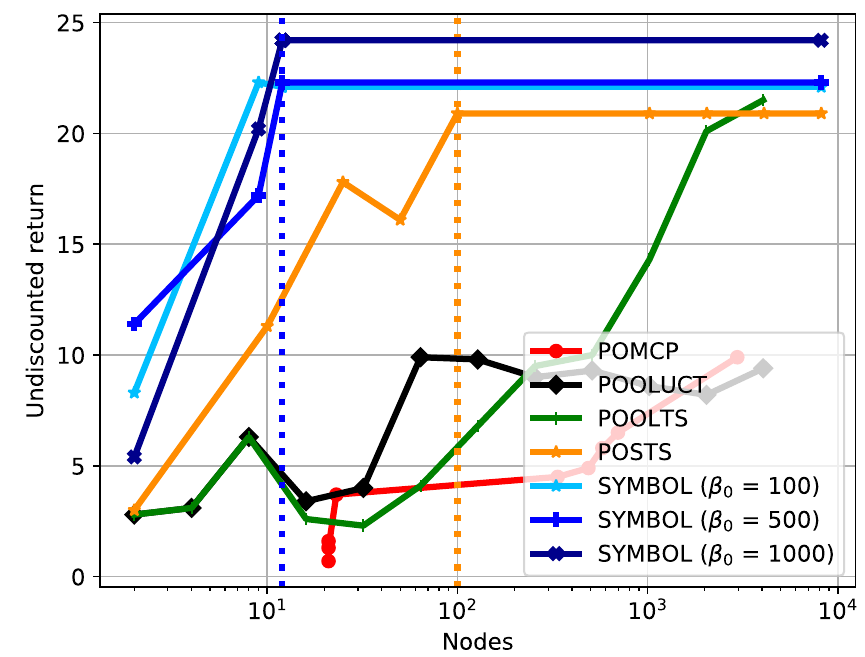}
     }
     \\
     \subfloat[\textit{Battleship}\label{fig:battleship_memory}]{%
       \includegraphics[width=0.23\textwidth]{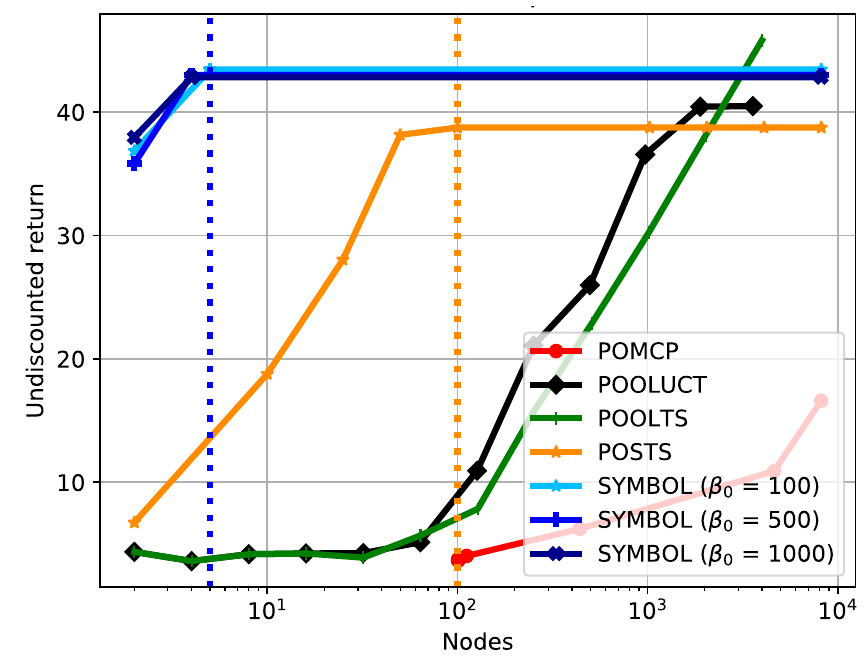}
     }
     \hfill
     \subfloat[\textit{PocMan}\label{fig:pocman_memory}]{%
       \includegraphics[width=0.23\textwidth]{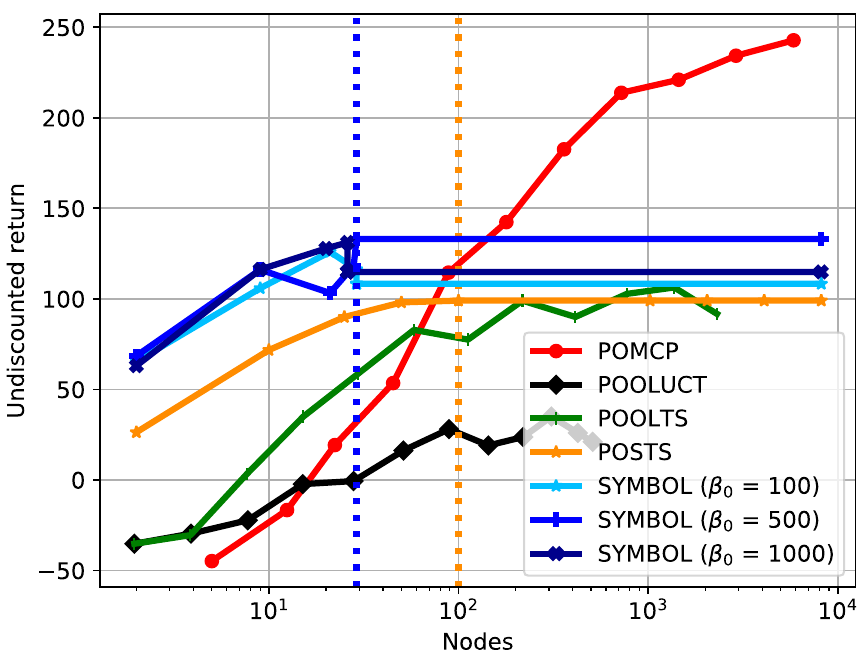}
     }
     \caption{Average performance of POMCP, POOLUCT, POOLTS, POSTS, and SYMBOL with memory bounds, $\textit{nb} = 4096$, $T = 100$, $\kappa = 8$, and $\epsilon = 6.4$. The vertical dotted lines indicate the maximum number of MABs used by SYMBOL (blue) and POSTS (orange).}
     \label{fig:symbol_performance_memory_tradeoff}
\end{figure}

We evaluated the performance-memory tradeoff of all approaches by introducing a memory capacity $\textit{nMEM}$, where the computation is interrupted, when the number of nodes exceeds $\textit{nMEM}$. For POMCP, we count the number of \textit{o-nodes} and \textit{a-nodes} (Fig. \ref{fig:closed_loop_planning}). For POOLTS and POOLUCT, we count the number of history distribution nodes (Fig. \ref{fig:open_loop_planning}). For SYMBOL and POSTS, we count $\textit{nMAB}$. POSTS always uses a planning horizon of $\textit{min}(T,\textit{nMEM})$ to satisfy the memory bound. The results are shown in Fig. \ref{fig:symbol_performance_memory_tradeoff} for $\textit{nb} = 4096$, $T = 100$, $\beta_{0} \in \{100,500,1000\}$ for POOLTS, POSTS, and SYMBOL, $\epsilon = 6.4$, and $\kappa = 8$.

In \textit{RockSample} and \textit{Battleship}, POMCP is outperformed by SYMBOL and POOLTS. SYMBOL always performs best in these domains, when $\textit{nMEM} \leq 1000$. POMCP performs best in \textit{PocMan} by outperforming SYMBOL, when $\textit{nMEM} > 100$ and POOLTS keeps up with SYMBOL, when $\textit{nMEM} > 1000$. SYMBOL always outperforms POSTS, while using a lower maximum number of MABs. POSTS is only able to keep up with the best SYMBOL setting in \textit{RockSample(11,11)} after creating 100 MABs, while SYMBOL only uses about 20 MABs for planning. POOLUCT performs worst except in \textit{Battleship}, improving less and slowest with increasing $\textit{nMEM}$.

\section{Discussion}
We presented SYMBOL, a general memory bounded approach to partially observable open-loop planning with an adaptive stack of Thompson Sampling bandits.

Our experiments show that SYMBOL is a good alternative to tree-based planning in POMDPs. SYMBOL is competitive against tree-based open-loop planning like POOLUCT and POOLTS and is able to keep up with POMCP in domains with large action spaces and low stochasticity like \textit{RockSample} or \textit{Battleship}. SYMBOL is robust w.r.t. the choice of the hyperparameters $\beta_{0}$ and $\epsilon$ in terms of performance, with $\epsilon$ strongly affecting the memory consumption in domains with high stochasticity as shown for \textit{PocMan} in Fig. \ref{fig:pocman_epsilon_memory}. $\epsilon$ should be sufficiently small to ensure stable convergence of the MABs, although more computation budget will be required to build up adequate MAB stacks, if $\epsilon$ is too small. Fig. \ref{fig:symbol_epsilon_sensitivity}e-h indicate that appropriate MAB stack sizes are highly domain dependent and cannot be generally specified beforehand without expensive parameter tuning. Thus, adaptive and robust approaches like SYMBOL seem to be promising for general and efficient decision making in POMDPs.

When restricting the memory capacity, SYMBOL clearly outperforms all tree-based approaches, while requiring significantly less MABs. Although being bounded by $T = 100$ at most, SYMBOL always created much less MABs in all domains, resulting in extremely memory efficient planning (Fig. \ref{fig:symbol_performance_memory_tradeoff}). POOLTS requires thousands of nodes to keep up with SYMBOL, while POMCP is only able to outperform SYMBOL in \textit{PocMan} after creating more than 100 nodes, which still consumes much more memory than SYMBOL (Fig. \ref{fig:pocman_memory}).

SYMBOL is able to outperform the fixed size stack approach POSTS, showing the effectiveness of the adaptive stack concept, where proper convergence is ensured by the convergence threshold $\epsilon$ and the convergence tolerance $\kappa$.

While state-of-the-art approaches to efficient online planning \cite{silver2010monte,somani2013despot,bai2014thompson} heavily rely on sufficient memory resources in highly complex domains, SYMBOL is a memory bounded alternative, which maintains an adaptive stack of MABs. SYMBOL is able to automatically adapt its stack according to the underlying domain without any prior domain knowledge. 
 
In the future, we plan to integrate SYMBOL into hierarchical planning to optimize macro-actions for certain subgoals.

\bibliographystyle{named}
\bibliography{references}

\end{document}